\documentclass{article}
\pdfoutput=1
\usepackage{graphicx} 

\usepackage{natbib}

\usepackage{algorithm}
\usepackage{algorithmic}
\usepackage{array}
\usepackage{amsmath}

\begin{document}

\title{Deep Embedding for Spatial Role Labeling}

\author{Oswaldo Ludwig, Xiao Liu, Parisa Kordjamshidi, Marie-Francine Moens}


\maketitle

\begin{abstract} 
This paper introduces the visually informed embedding of word (VIEW), a continuous vector representation for a word extracted from a deep neural model trained using the Microsoft COCO data set to forecast the spatial arrangements between visual objects, given a textual description. The model is composed of a deep multilayer perceptron (MLP) stacked on the top of a Long Short Term Memory (LSTM) network, the latter being preceded by an embedding layer. The VIEW is applied to transferring multimodal background knowledge to Spatial Role Labeling (SpRL) algorithms, which recognize spatial relations between objects mentioned in the text. This work also contributes with a new method to select complementary features and a fine-tuning method for MLP that improves the $F1$ measure in classifying the words into spatial roles. The VIEW is evaluated with the Task 3 of SemEval-2013 benchmark data set, SpaceEval.
\end{abstract}

\section{Introduction}
\label{intro}


One of the essential functions of natural language is to describe location and translocation of objects in space. Spatial language can convey complex spatial relations along with polysemy and ambiguity inherited in natural language. Therefore, a formal spatial model is required, to focus on some particular spatial aspects. This paper address a layer of linguistic conceptual representation, called spatial role labeling (SpRL), which predicts the existence of spatial information at the sentence level by identifying the words that play a particular spatial role as well as their spatial relationship \cite{kordjamshidi2011spatial}.

An issue in extracting spatial semantics from natural language is the lack of annotated data on which machine learning can be employed to learn and extract the spatial relations. Current SpRL algorithms rely strongly on feature engineering, which has the advantage of encoding human knowledge, thus compensating for the lack of annotated training data. This work preserves the previous contributions on feature engineering of \cite{kordjamshidi2011spatial} while adding a new set of features learned from multimodal data, i.e. the visually informed embedding of word (VIEW). 

Multimodal data is usually associated with multimodal representation learning, which has been studied by various authors, such as \cite{srivastava2012multimodal}, which uses deep Boltzmann machines for representing joint multimodal probability distributions over images and sentences, \cite{karpathy2014deep}, which then embed fragments of images (objects) and fragments of sentences (dependency tree relations) into a common space for bidirectional retrieval of images and sentences, and \cite{kiros2014unifying}, which unify joint image-text embedding models with multimodal neural language models to rank images and sentences (as well as to generate descriptions for images) using a long short term memory (LSTM) network to process text and deep convolutional network (CNN) to process images.

Similar to the work \cite{kiros2014unifying} our model learns embedding from multimodal data and applies LSTM to process textual information. However, unlike most of the works on multimodal representation learning, which jointly map visual and textual information into a common embedding space, our work aims at providing embeddings only for words, but encoding spatial information extracted from the image annotations. The idea is to learn VIEW by pipelining an embedding layer into a deep architecture trained by back propagation to predict the spatial arrangement between the visual objects annotated in the pictures, given the respective textual descriptions. In this sense, unlike \cite{karpathy2014deep} and \cite{kiros2014unifying}, we don't need a CNN, because the spatial information which is relevant for our purpose is provided directly by the position of the bounding boxes containing the visual objects annotated in the images, as detailed in Section \ref{Sec_model}. Our VIEW is used as a vehicle to transfer spatial information from multimodal data to SpRL algorithms.


The paper is organized as follows. Section \ref{Sec_model}, we describe the model setting, from the annotation style through the modeling of the deep neural network, whose training algorithm is described in Section \ref{training}. Section \ref{SpRL} describes how the spatial embedding is applied in SpRL, as well as the algorithm developed to select the best complementary embedding features and the fine-tuning method able to deal with the tradeoff of precision and recall, aiming at the largest $F1$. Section \ref{experiments} reports and discusses the experiments, while Section \ref{conclusion} summarizes the major findings.

\section{Problem definition and research questions}
\label{task}

The SpRL algorithm recognizes spatial objects in language (i.e. trajector and landmark) and their spatial relation signaled by the spatial indicator. The trajector is a spatial role label assigned to a word or a phrase that denotes an object of a spatial scene, more specifically, an object that moves. A landmark is a spatial role label assigned to a word or a phrase that denotes the location of this trajector object in the spatial scene. The spatial indicator is a spatial role label assigned to a word or a phrase that signals the spatial relation trajector and landmark.

In this work we apply the SpRL algorithm developed for the work \cite{kordjamshidi2011spatial}, which models this problem as a structured prediction task \cite{taskar2005learning}, that is, it jointly recognizes the spatial relation and its composing role elements in text. The SpRL algorithm receives as input a natural language sentence, such as ``There is a white large statue with spread arms on a hill'', having a number of words, in this case identified as $\textbf{w}=\left\{\textbf{w}_1,\ldots,\textbf{w}_{12}\right\}$, where $\textbf{w}_i$ is the $i^{th}$ word in the sentence.

Each word in the sentence that can be part of a spatial relation (e.g., nouns, prepositions) is described by a vector of the local features denoted by $\phi_{word}(\textbf{w}_i)$, including linguistically motivated lexical, syntactical and semantical features of words, such as the lexical surface form, its semantic role, its part-of-speech and the lexical surface form of words in the neighborhood. This feature vector is used to relate a word with its spatial role, i.e. spatial indicator, trajector or landmark, hence further represented by $sp$, $tr$ and $lm$, respectively.

There are also descriptive vectors of pairs of words, referred to as $\phi_{pair}(\textbf{w}_i,\textbf{w}_j)$, encoding the linguistically motivated features of pairs of words and their relational features, such as their relative position in the sentence, their distance in terms of number of words and the path between them obtained with a syntactic parser. The SpRL model is trained on a training set of sentences annotated with the above output labels. Following training, the system outputs all spatial relations found in a sentence of a test set composed of the sentences and their corresponding spatial roles.

The main research question approached in this paper regards the possibility of improving the quality of $\phi_{word}(\textbf{w}_i)$ by concatenating the VIEW in this feature vector. It derives a secondary research question on the possibility of encoding visual information from COCO images into the word embeddings by learning a model able to map from the captions to a simplified representation of the visual objects annotated in the corresponding image and their relative position. We assume that the necessary condition to correctly forecast the visual output, given the textual description, is that the embedding layer (which is in the model pipeline) is successfully encoding the spatial information of the textual description, assuring a suitable word embedding for this specific task related with SpRL.

Another research question regards the importance of feature selection in order to discard embedding features that are not directly related to the SpRL task, since our data set derived from COCO is not created for SpRL; therefore, some features can act as noise for SpRL.

\section{The model setting}
\label{Sec_model}

\begin{figure}[ht]
\vskip 0.0in
\begin{center}
\centerline{\includegraphics[width=\columnwidth]{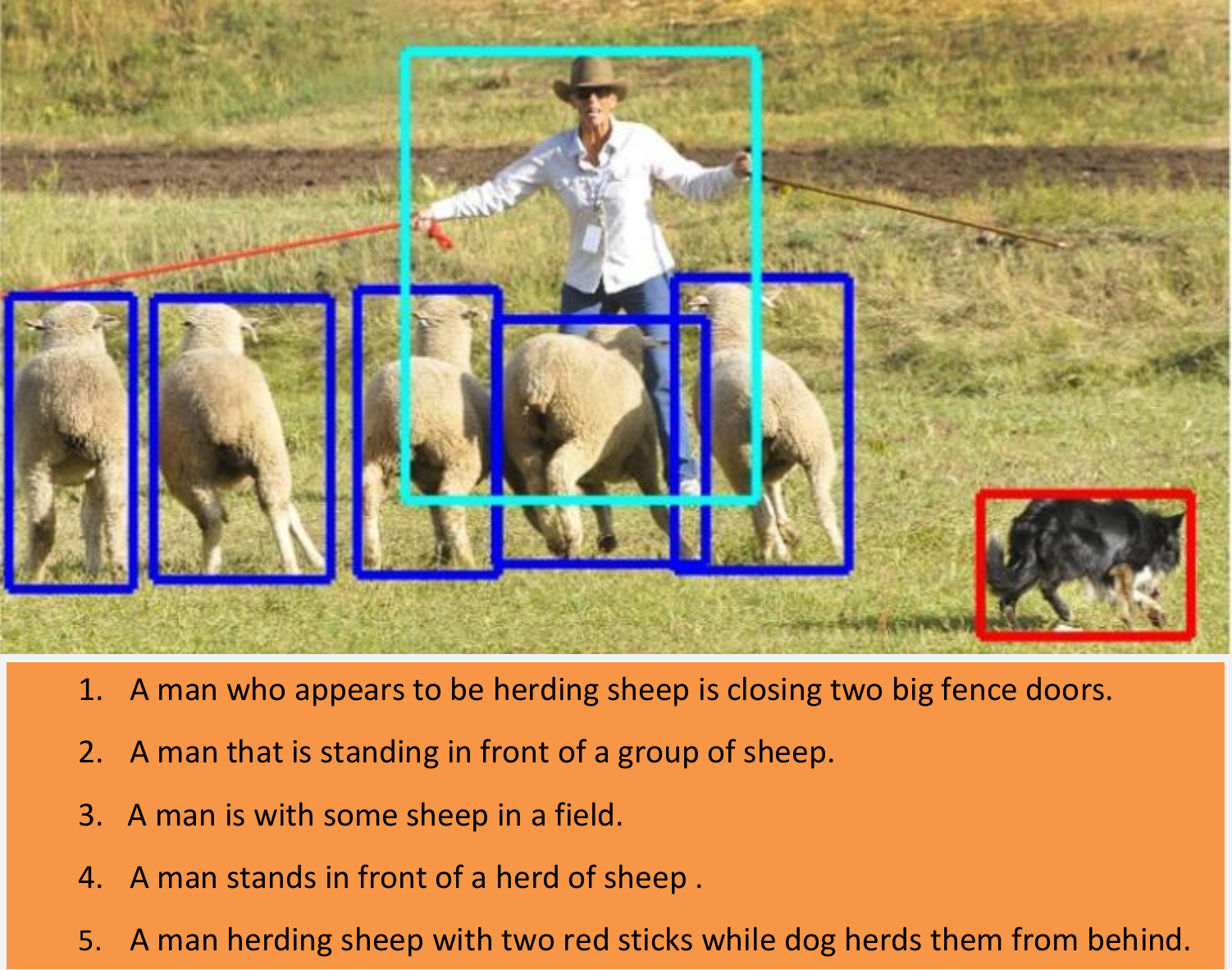}}
\caption{This figure shows an image from COCO with annotated bounding boxes and captions \cite{lin2014microsoft}.}
\label{Fig_COCO}
\end{center}
\vskip -0.3in
\end{figure}

The Microsoft COCO data set \cite{lin2014microsoft} is a collection of images featuring complex everyday scenes which contain common visual objects in their natural context. COCO contains photos of 91 object types with a total of 2.5 million labeled instances in 328k images. Each image has five written caption descriptions. The visual objects within the image are tight-fitted by bounding boxes from annotated segmentation masks, as can be seen in Fig.\ref{Fig_COCO}. 


\begin{figure}[ht]
\vskip -0.1in
\begin{center}
\centerline{\includegraphics[width=\columnwidth]{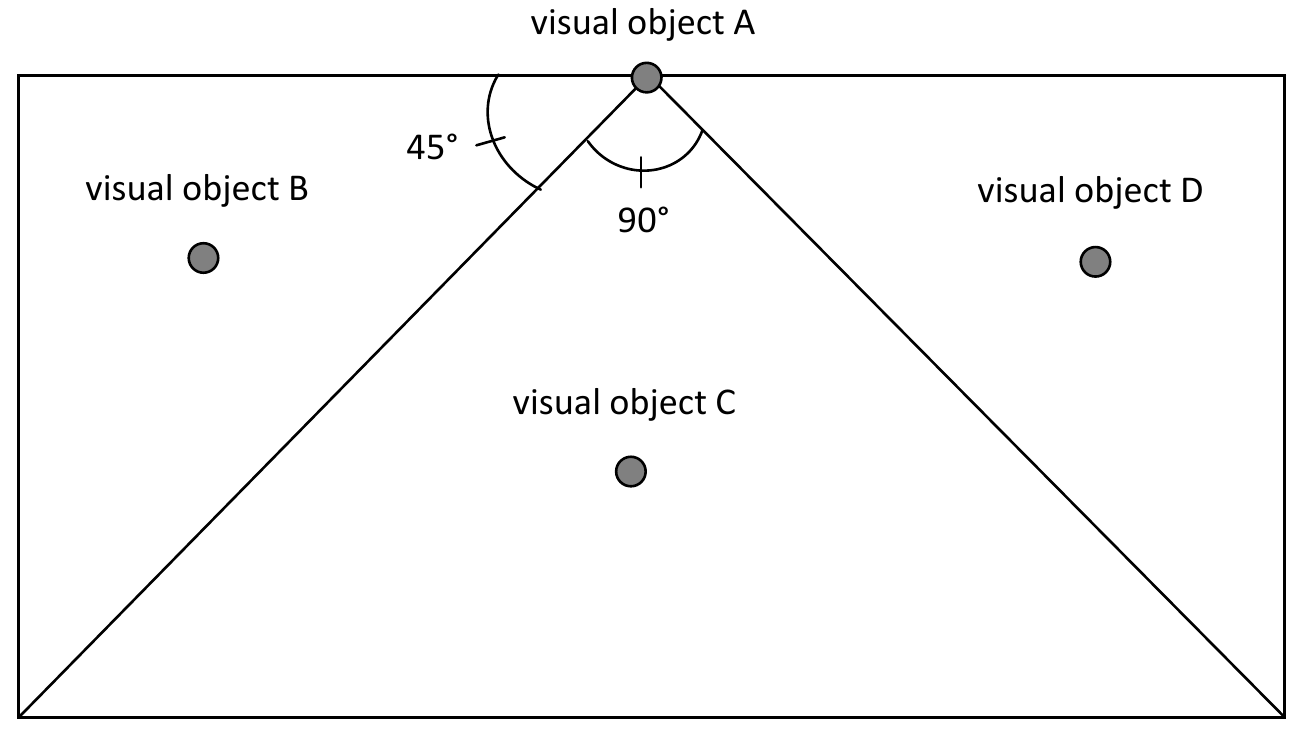}}
\caption{This figure exemplifies the center of four bounding boxes containing the visual objects A, B, C and D, whose spatial annotations yield triplets such as $below(C, A)$, $beside(B, A)$, $beside(D, A)$.}
\label{fig_spatial}
\end{center}
\vskip -0.3in
\end{figure}

Our annotation system automatically derives a less specific spatial annotation about the relative position between the center points of the bounding boxes containing visual objects, given COCO's annotation of the coordinates of the bounding boxes. Our annotation style is ruled by the predicates $alone(v_1)$, $below(v_1,v_2)$ and $beside(v_1,v_2)$, where $v_1$ and $v_2$ are visual objects, see Fig.\ref{fig_spatial}. This information is encoded in a sparse target vector, $Y$, where the first three positions encode the predicate in one-hot vector style, and the ensuing positions encode the index of the visual objects, also in one-hot vector style, i.e. 3 positions to encode the predicates plus 91 positions to encode the index of the first argument (visual object) plus other 91 positions for the second argument, totalizing 185 positions in the target vector. If the predicate is $alone(v_1)$, i.e. when a single visual object is annotated in the image, the last 91 positions are all zeros.


Despite having a large number of annotated objects per image, MS-COCO has several objects belonging to the same category per image, and so, a small number of categories per image. On average MS-COCO data set contains 3.5 categories per image, yielding 7.7 instances per image (see Section 5 of \cite{lin2014microsoft}). Our annotation system is based on two assumptions: 1) learning the spatial arrangement between visual objects belonging to the same category is not useful: 2) objects placed at the center of the image are more salient, and so, they are
likely to be present in the captions. Therefore, our system ranks the bounding
boxes from the most centered box to the least centered box and starts by selecting
the most centered bounding box as the first visual object. Then, it searches from
the second higher ranked bounding box to those with a lower rank, until it reaches
a visual object belonging to a different category or the end of the list of
instances, i.e. our system selects only a pair of visual objects belonging to
different categories.

In summary, the system learns how to map the captions to the spatial relation between the most salient pair of objects belonging to different categories. For instance, in Fig.\ref{Fig_COCO} the system selects the man and the most centered sheep (not the dog in the corner of the image) to generate the target output for all the five captions, yielding five training examples. Notice that the dog is only cited in one of the five captions.

\subsection{The embedding model}
\label{model1}

The input of our deep model is the textual information provided by COCO's captions, i.e., a sequence\footnote{The words, encoded in one-hot vectors, are provided sequentially to the model.} of words encoded in a set of $K$-dimensional one-hot vectors, where $K$ is the vocabulary size, here assumed as 8000 words. Since COCO's captions are shorter than 30 words, our system cuts texts after the limit of 30 words; therefore, the data pair for the $i^{th}$ caption is composed by a sparse matrix $X_i$ of dimension $8000\times30$ per target vector, $y_i$, of 185 dimensions, as explained in the previous paragraph. 
Captions with less than 30 words are encoded into a matrix $X_i$ whose the first columns are filled with all-zeros. Each figure has five associated captions, which yields five training examples with the same target vector, but different input vectors.

Our deep model is trained to forecast the visual objects and their spatial relation (i.e. $alone(v_1)$, $below(v_1,v_2)$ and $beside(v_1,v_2)$), given the textual description in the caption.

\begin{figure}[ht]
\begin{center}
\centerline{\includegraphics[width=\columnwidth]{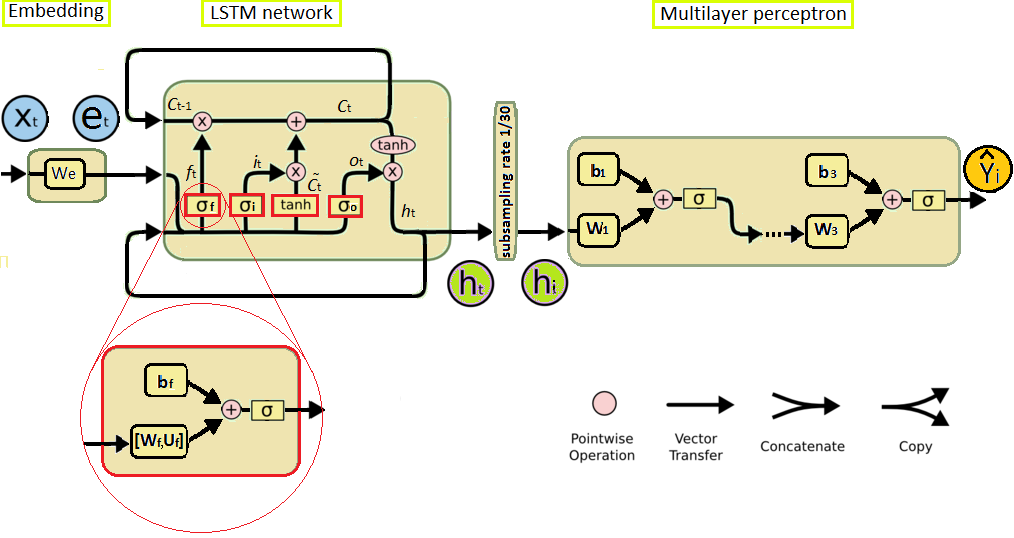}}
\caption{The deep neural model, where red border boxes represent sets of neurons, as exemplified for the box $\sigma_f$.}
\label{Fig_net}
\end{center}
\end{figure}

The model was implemented in Keras\footnote{http://keras.io/} and it is pipeline composed of a linear embedding layer, an original version of LSTM network, as proposed in \cite{hochreiter1997long}, and a deep multilayer perceptron (MLP), see Fig.\ref{Fig_net}. The embedding layer receives the sparse input $X_i$, representing a sentence, and encodes the one-hot vector representation of the words (i.e. the columns of $X_i$) into a set of $30$ dense vectors of $N_w$ dimensions that are provided sequentially to the LSTM, which extracts a $N_s$-dimensional vector from the sentence (after 30 iterations). This vector representation of the sentence (i.e. a sentence-level embedding) is mapped to the sparse spatial representation $y_i$ by the MLP.


We also evaluated an architecture where the LSTM directly predicts the sparse output vector (without MLP), but the result was better using the MLP. The latent representation of sentences, i.e. the sentence-level embedding, makes possible the choice of a suitable dimension for the LSTM output, rather than forcing the LSTM output to be the sparse 185-dimensional target vector. 


The $i^{th}$ caption yields a matrix $X_i$ from which a set of $30$ input vectors, here represented by $x_t$ ($t=1,\ldots,30$), are sequentially given to the model.

Our model starts by computing the word embedding:
\begin{equation}
\label{model}
e_t = W_e x_t
\end{equation}
where the adjustable embedding matrix, $W_e$, has the dimension of $8000\times N_w$, i.e. it maps from the 8000-dimensional one-hot vector representation of words to a $N_w$-dimensional continuous vector representation $e_t$. Then the system calculates the state variables of the LSTM, starting by the input gate $i_t$ and the candidate value for the states of the memory cells $\widetilde{C_t}$ at iteration $t$, as follows:
\begin{equation}
\label{modela}
i_t = \sigma(W_i e_t + U_i h_{t-1} + b_i)
\end{equation}
\begin{equation}
\label{modelb}
\widetilde{C_t} = tanh(W_c e_t + U_c h_{t-1} + b_c)
\end{equation}
where $\sigma(\cdot)$ represents the sigmoid function, $W_i$, $U_i$, $W_c$, $U_c$, $b_i$ and $b_c$ are adjustable parameters and $h_{t-1}$ is the LSTM output at the previous iteration. Having $i_t$ and $\widetilde{C_t}$ the system can compute the activation of the memory forget gates, $f_t$, at iteration $t$:
\begin{equation}
\label{modelc}
f_t = \sigma(W_f e_t + U_f h_{t-1} + b_f)
\end{equation}
where $W_f$, $U_f$ and $b_f$ are adjustable parameters. Having $i_t$, $f_t$ and the candidate state value $\widetilde{C_t}$, the system can compute the new state of the memory cells, $C_t$, at iteration $t$:
\begin{equation}
\label{modeld}
C_t = i_t * \widetilde{C_t} + f_t * C_{t-1}
\end{equation}
where $\ast$ represents the point-wise multiplication operation. With the new state of the memory cells, $C_t$, the system can compute the value of their output gates, $o_t$:
\begin{equation}
\label{modele}
o_t = \sigma(W_o e_t + U_o h_{t-1} + b_o)
\end{equation}
were $W_o$, $U_o$ and $b_o$ are adjustable parameters. Having $o_t$, the system can finally calculate the output of the LSTM, $h_t$:
\begin{equation}
\label{modelf}
h_t = o_t * tanh(C_t)
\end{equation}

At this point we have both word-level and sentence-level embeddings, given by (\ref{model}) and (\ref{modelf}), respectively. Each $N_s$-dimensional sentence embedding is produced after 30 iteration of LSTM (the adopted sentence length), when it is ready to be processed by the MLP. Therefore, the LSTM output is sub-sampled in the rate of $1/30$, yielding $h_i$, i.e. the input for the MLP (note the changing of the index variable in relation to (\ref{modelf})).  

The adopted MLP has two sigmoidal hidden layers and a sigmoidal output layer. The optimal number of hidden layers was empirically determined based on the performance on MS-COCO. The MLP model is given by:
\begin{equation}
\label{modelMLP}
\begin{array}{l}
      yh_{(1,i)}=\sigma\left(W_{1_{ } } h_i+b_{1}\right) \\
		  yh_{(2,i)}=\sigma\left(W_{2_{ } } yh_{(1,i)}+b_{2}\right) \\
      \hat{y}_i=\sigma\left(W_{3_{ } } yh_{(2,i)}+b_{3}\right)
\end{array}
\end{equation}
where $W_{j}$ and $b_{j}$ are the weight matrix and bias vector of layer $j$, respectively, and $yh_{(j,i)}$ is the output vector of the hidden layer $j$.




\section{Model Training}
\label{training}

After evaluating different objective functions combined with different activity and weight regularizers available in Keras, we decided to implement a custom objective function, gathering some ideas from the support vector learning. Our training method yields the following constrained optimization problem:
\begin{equation}
\label{optimization}
\min_{W_e,\theta_l,\theta_m} \frac{1}{N_e N_o}\sum_{i=1}^{N_e}\sum_{j=1}^{N_o}\max\left(1-\hat{y}_{(i,j)}\left(2y_{(i,j)}-1\right),0\right)
\end{equation}
subject to:
\begin{equation}
\label{constrain1}
\left\|W_l^{neu}\right\| \leq 1,_{}  \left.\right.^{l=1,2,3}_{\forall neu}
\end{equation}
where the objective function (\ref{optimization}) is the Hinge loss with the $j^{th}$ position of the intended output vector for the $i^{th}$ caption, $y_{(i,j)}\in\left\{0,1\right\}$, scaled and shifted to assume the values $-1$ or $1$, $N_e$ is the cardinality of the training data set, $N_o=185$ is the dimension of the output vector, $W_e$ is the weight matrix of the embedding, $\theta_l=\left\{W_i,W_f,W_c,W_o,U_i,U_f,U_c,U_o,b_i,b_f,b_c,b_o\right\}$ is the set of adjustable LSTM parameters, $\theta_m=\left\{W_1,W_2,W_3,b_1,b_2,b_3\right\}$ is the set of adjustable MLP parameters, $\hat{y}_{(i,j)}$ is the $j^{th}$ position of the output vector estimated by our model for the $i^{th}$ caption and $W_l^{neu}$ is the vector of synaptic weights of the neuron $neu$ of the layer $l$.

Keras allows to set constraints on network parameters during optimization. The adopted set of constraints (\ref{constrain1}) regularizes the model by upper bounding the norm of the vector of synaptic weights of the MLP neurons. Note that the adopted loss function (\ref{optimization}) only penalizes examples that violate a given margin or are misclassified, i.e. an estimated output smaller than 1 in response to a positive example or an estimated output larger than -1 in response to a negative example (these training examples can be understood as support vectors). The other training examples are ignored during the optimization, i.e. they don't participate in defining the decision surface.


The best optimization algorithm for our model was Adam \cite{kingma2014adam}, an algorithm for first-order gradient-based optimization based on adaptive estimates of lower-order moments. 


\section{Applying the spatial-specific embedding in SpRL}
\label{SpRL}

We apply VIEW in SpRL by simply concatenating it with the original feature vector, $\phi_{word}(\textbf{w}_i)$, generated by the SpRL algorithm \cite{kordjamshidi2011spatial} for the words that are candidate for $sp$, $tr$ and $lm$.

\subsection{Selecting complementary Features}
\label{feature}

Our aim is to select complementary features from the VIEW so as to maximize the mutual information between the target output, here represented by the scalar random variable $r$, and the selected features, represented by the random variables $x_1 \in\mathcal{X}_1,\ldots, x_n \in\mathcal{X}_n$, while minimizing the mutual information between the selected features and the original SpRL features, $\phi_{word}(\cdot)$.

The method introduced in this section requires a scalar random variable, $r$, as target output. However, the target output of SemEval is a 3-dimensional one-hot vector indicating the spatial roles, i.e. $sp$, $tr$ and $lm$. Therefore, we convert this binary number with 3 binary digits into a decimal number, i.e. a scalar.  

The mutual information is given by:
\begin{equation}
\label{MI}
\begin{array}{l}
I(x_1,\ldots, x_n;r) =\\
H(x_1,\ldots, x_n)+H(r)-H(x_1,\ldots, x_n,r)
\end{array}
\end{equation}
where $n$ is the arbitrary number of selected features and $H(.)$ represents the entropy of a set of random variables, given by:
\begin{equation}
\label{MI2}
\begin{array}{l}
H(x_1, \ldots, x_n) =\\
-\int_{\mathcal{X}_1}\ldots\int_{\mathcal{X}_n} p(x_1, \ldots, x_n)\log p(x_1, \ldots, x_n)dx_1 \ldots dx_n
\end{array}
\end{equation}
Even assuming a discrete approximation for the joint density, $p(x_1, \ldots, x_n)$, e.g. normalized histograms, the calculation of (\ref{MI2}) for several random variables is computationally unfeasible. Therefore, we adopt an indirect approach by applying the principle of Max-Relevance and Min-Redundancy \cite{Peng}. According to this principle it is possible to maximize (\ref{MI2}) by jointly solving the following two problems:
\begin{equation}
\label{MRelmRed}
\max_{i_1,\ldots,i_n}\phi(i_1, \ldots, i_n)
\end{equation}
where $i_n$ is the index of the $n^{th}$ selected feature, $\phi(i_1, \ldots, i_n)=V(i_1, \ldots, i_n)-D(i_1, \ldots, i_n)$,
\begin{equation}
\label{rel}
V(i_1, \ldots, i_n)=\frac{1}{n}\sum_{k=1}^n I(x_{i_k};r),
\end{equation}
\begin{equation}
\label{red}
D(i_1, \ldots, i_n)=\frac{1}{n^2}\sum_{j=1}^n\sum_{k=1}^n I(x_{i_j};x_{i_k})
\end{equation}

The idea is to find the set of indexes, $i_1, \ldots, i_n$, that simultaneously maximize the relevance (\ref{rel}) and minimize the redundancy (\ref{red}). Notice that this procedure requires the calculation of a matrix $S \in R^{n\times n}$ whose the elements are the mutual information values $I(x_{i_j};x_{i_k})$. However, this is a naive approach, since this method doesn't take into account the redundancy between the embedding features and the original 8099 SpRL features of $\phi_{word}(\cdot)$. The computational cost increases significantly by considering the whole problem. Let $x_k^{SpRL}$ be the $k^{th}$ original feature from $\phi_{word}(\cdot)$, then the complete problem can be modeled as:
\begin{equation}
\label{MRelmRed2}
\max_{i_1,\ldots,i_n}\phi(i_1, \ldots, i_n)
\end{equation}
\begin{align}
\label{rel2}
V(i_1, \ldots, i_n)=\frac{1}{n+m}\left(\sum_{k=1}^n I(x_{i_k};r)+\right. \nonumber \\
\left.\sum_{j=1}^m I(x_{i_j}^{SpRL};r)\right),
\end{align}
\begin{align}
\label{red2}
D(i_1, \ldots, i_n)=\frac{1}{n^2+nm+m^2}\left(\sum_{j=1}^n\sum_{k=1}^n I(x_{i_j};x_{i_k})+\right. \nonumber \\
\left.\sum_{l=1}^n\sum_{z=1}^m I(x_{i_l};x_{i_z}^{SpRL})+\sum_{w=1}^m\sum_{q=1}^m I(x_{i_w}^{SpRL};x_{i_q}^{SpRL})\right)
\end{align} 
where $m$ is the number of original features from $\phi_{word}(\cdot)$. Fortunately, the terms $\sum_{j=1}^m I(x_{i_j}^{SpRL};y)$ and $\sum_{w=1}^m\sum_{q=1}^m I(x_{i_w}^{SpRL};x_{i_q}^{SpRL})$ are constant in relation to the manipulated indexes, i.e. the mutual information between the original SpRL features and the output, as well as the mutual information between pairs of SpRL features, don't matter for this optimization problem, alleviating the computational cost. Therefore, (\ref{MRelmRed2})-(\ref{red2}) can be simplified as follows:
\begin{equation}
\label{MRelmRed3}
\max_{i_1,\ldots,i_n}\phi(i_1, \ldots, i_n)
\end{equation}
\begin{equation}
\label{rel3}
V(i_1, \ldots, i_n)=\frac{1}{n}\sum_{k=1}^n I(x_{i_k};r),
\end{equation}
\begin{align}
\label{red3}
D(i_1, \ldots, i_n)=\frac{1}{n^2+nm}\left(\sum_{j=1}^n\sum_{k=1}^n I(x_{i_j};x_{i_k})+\right. \nonumber \\
\left.\sum_{l=1}^n\sum_{z=1}^m I(x_{i_l};x_{i_z}^{SpRL})\right)
\end{align}

The optimization problem (\ref{MRelmRed3})-(\ref{red3}) requires the calculation of a matrix with the pairwise mutual information values of dimension $n\times m$, in the place of the $m\times m$ matrix required by (\ref{red2}). In our case it means a computational effort around 100 times smaller.

We solved (\ref{MRelmRed3})-(\ref{red3}) by slightly adapting the Feature Selector based on Genetic Algorithm and Information Theory\footnote{http://www.mathworks.com/matlabcentral/} \cite{ludwig_NN}.

\subsection{Maximizing the $F1$}
\label{max_F1}

After having the selected features from the embedding concatenated with the original SpRL features, we train an MLP on SemEval annotated data to predict the spatial role of the words. The adopted MLP has a single sigmoid hidden layer and a linear output layer.

One of the issues that we observe in applying MLP trained with MSE on SpRL data is the unbalanced relation between precision and recall that worsens with the use of the embedding, resulting in damage on the $F1$. This issue is usually solved by manipulating the threshold; however, a larger gain on $F1$ can be obtained by manipulating all the parameters of the output layer. Therefore, we propose a fine-tuning of the output layer, by maximizing an approximation of $F1$ squared. We start by analyzing the simplest approach:
\begin{equation}
\label{maxF1}
\max_{w,b} F1^2
\end{equation}
where $w$ and $b$ are the adjustable parameters of the linear output layer of the MLP. $F1$ is function of the true positive $(TP)$ and true negative examples $(TN)$, as follows:
\begin{equation}
\label{F1}
F1=\frac{2TP}{N+TP-TN}
\end{equation}
\begin{equation}
\label{TP}
TP=\frac{1}{2}\sum_{i=1}^{N_p}\left(1+\varphi(w_{ }x_i^p+b)\right)
\end{equation}
\begin{equation}
\label{TN}
TN=\frac{1}{2}\sum_{j=1}^{N_n}\left(1-\varphi(w_{ }x_j^n+b)\right)
\end{equation}
where $\varphi(\cdot)= sign(\cdot)$ returns $1$ or $-1$ according to the sign of the argument, $N$ is the number of examples, $N_p$ and $N_n$ are the number of positive and negative examples, respectively, and $x_i^p$ and $x_j^n$ are the outputs of the hidden layer of the MLP (i.e. the inputs of the output layer) for the $i^{th}$ positive example and the $j^{th}$ negative example, respectively.

Unfortunately, the $sign$ function is not suitable for gradient based optimization methods; therefore, we approximate this function by hyperbolic tangent, yielding the approximate $\tilde{TP}$, $\tilde{TN}$ and $\tilde{F1}$, whose relation is given by:
\begin{equation}
\label{approxF1}
\tilde{F1}=\frac{2\tilde{TP}}{N+\tilde{TP}-\tilde{TN}}
\end{equation}

To derive a lower bound on $F1$ as a function of $\tilde{F1}$, $\tilde{TP}$ and $\tilde{TN}$, let us analyze the bounds on the difference $\left(sign(\cdot)-tanh(\cdot)\right)$ by one-sided limit:
\begin{equation}
\label{lim_TP_neg}
\lim_{v\rightarrow 0^{-}} \left(sign(v)-tanh(v)\right)=-1
\end{equation}
\begin{equation}
\label{lim_TP_pos}
\lim_{v\rightarrow 0^{+}} \left(sign(v)-tanh(v)\right)=1
\end{equation}
From (\ref{lim_TP_neg}), (\ref{lim_TP_pos}) and (\ref{TP}) we can derive bounds on $TP$ as a function of its approximation $\tilde{TP}$:
\begin{equation}
\label{bounds_TP}
\tilde{TP}-\frac{N_p}{2}<TP<\tilde{TP}+\frac{N_p}{2}
\end{equation}
Notice that we can relax these bounds by substituting $N_p$ by the largest value between $N_p$ and $N_n$, henceforward called $N_l$, i.e.:
\begin{equation}
\label{bounds_TP_relaxed}
\tilde{TP}-\frac{N_l}{2}<TP<\tilde{TP}+\frac{N_l}{2}
\end{equation}
Similar bounds can be derived for $TN$:
\begin{equation}
\label{bounds_TN_relaxed}
\tilde{TN}-\frac{N_l}{2}<TN<\tilde{TN}+\frac{N_l}{2}
\end{equation}
By substituting the lower bounds of (\ref{bounds_TP_relaxed}) and (\ref{bounds_TN_relaxed}) into (\ref{F1}) and using (\ref{approxF1}) we can derive a lower bound on $F1$:
\begin{equation}
\label{alpha}
F1>(1-\frac{N_l}{2\tilde{TP}})\tilde{F1}
\end{equation}   
From (\ref{alpha}) we conclude that it is desirable to have $(1-\frac{N_l}{2\tilde{TP}})>0$, since this is a necessary condition (but not sufficient) to have the lower bound of $F1$ increasing together with $\tilde{F1}$. Therefore, we propose the constrained optimization problem:
\begin{equation}
\label{maxF1approx}
\begin{aligned}
&\max_{w,b,\zeta} \tilde{F1}^2+C\zeta\\
&\text{s.t.}
&\tilde{TP}>\frac{N_l}{2}+\zeta
\end{aligned}
\end{equation}
where $\zeta$ is a slack variable. To solve this optimization problem we need the following derivatives:
\begin{align}
\label{dF12dw}
\frac{\partial \tilde{F1}^2}{\partial w}=2\tilde{F1}\left(2\frac{\partial\tilde{TP}}{\partial w}\left(N+\tilde{TP}-\tilde{TN}\right)^{-1}\right. + \nonumber \\
\left.2\tilde{TP}\left(\tilde{TN}-\tilde{TP}-N\right)^{-2}\left(\frac{\partial\tilde{TN}}{\partial w}-\frac{\partial\tilde{TP}}{\partial w}\right)\right)
\end{align}
\begin{align}
\label{dF12db}
\frac{\partial \tilde{F1}^2}{\partial b}=2\tilde{F1}\left(2\frac{\partial\tilde{TP}}{\partial b}\left(N+\tilde{TP}-\tilde{TN}\right)^{-1}\right. + \nonumber \\
\left.2\tilde{TP}\left(\tilde{TN}-\tilde{TP}-N\right)^{-2}\left(\frac{\partial\tilde{TN}}{\partial b}-\frac{\partial\tilde{TP}}{\partial b}\right)\right)
\end{align}
\begin{equation}
\label{diffTPw}
\frac{\partial\tilde{TP}}{\partial w}=\frac{1}{2}\sum_{i=1}^{N_p}\left(tanh'(w_{}x_i^p+b)x_i^p\right)
\end{equation}
\begin{equation}
\label{diffTNw}
\frac{\partial\tilde{TN}}{\partial w}=-\frac{1}{2}\sum_{j=1}^{N_n}\left(tanh'(w_{}x_j^n+b)x_j^n\right)
\end{equation}
\begin{equation}
\label{diffTPb}
\frac{\partial\tilde{TP}}{\partial b}=\frac{1}{2}\sum_{i=1}^{N_p}\left(tanh'(w_{}x_i^p+b)\right)
\end{equation}
\begin{equation}
\label{diffTNb}
\frac{\partial\tilde{TN}}{\partial b}=-\frac{1}{2}\sum_{j=1}^{N_n}\left(tanh'(w_{}x_j^n+b)\right)
\end{equation}
where $tanh'(\cdot)=1-tanh^2(\cdot)$.

\section{Experimental Settings}
\label{experiments}

In this section our methods are evaluated by means of experiments in the SemEval-2013 benchmark data set, the Task 3, SpaceEval. We start by evaluating the embedding model on COCO, then we evaluate the contribution of the spatial-specific embedding in the multiclass classification of words into spatial roles and finally in the structured prediction of spatial triplets by using the algorithm of \cite{kordjamshidi2011spatial}.

\subsection{Evaluating the embedding model on COCO}
\label{eval_embedding}

This subsection reports the performance indexes of the deep model described in Section \ref{model1} in predicting our annotation on the COCO testing data. The model was trained on 135015 captions and evaluated on a test set composed by 67505 captions.

According to our experiments, the deep model has its best performance on the test data when having a 200-dimensional word embedding and a $300$-dimensional sentence embedding, meaning that the embedding matrix has dimension $8000\times200$ and the LSTM receives a 200-dimensional vector and outputs a 300-dimensional vector, i.e. $N_w=200$ and $N_s=300$. The best setup for the MLP is $300\times250\times200\times185$, i.e. with two sigmoidal hidden layers containing 250 and 200 neurons, respectively.

The performance indexes on the test data are provided for the model trained with mean squared error (MSE) and our Hinge-like multiclass loss (\ref{optimization}) for the sake of comparison, as shown in Table \ref{deep}.

\begin{table}[!htbp]
\centering
\caption{Results on test data using the model trained with MSE and the Hinge-like loss (\ref{optimization}).} 
\label{deep}
\begin{tabular}{lll}
\hline
class & acc (MSE) & acc (Hinge loss)   \\
\hline
$spatial$ $relation$ & 0.997 & 0.997 \\                      
$visual$ $object$ $\#1$ & 0.831 & 0.863 \\
$visual$ $object$ $\#2$ & 0.735 & 0.747 \\
\hline
\end{tabular}
\end{table}

As can be seen in Table \ref{deep}, the model has a better performance in predicting the $spatial$ $relation$, since it can assume only 3 discrete values, while the $visual$ $object$ can assume 91 discrete values. The Hinge-like multiclass loss presents slightly better performance indexes.

\begin{figure}[ht]
\vskip -0.1in
\begin{center}
\centerline{\includegraphics[width=\columnwidth]{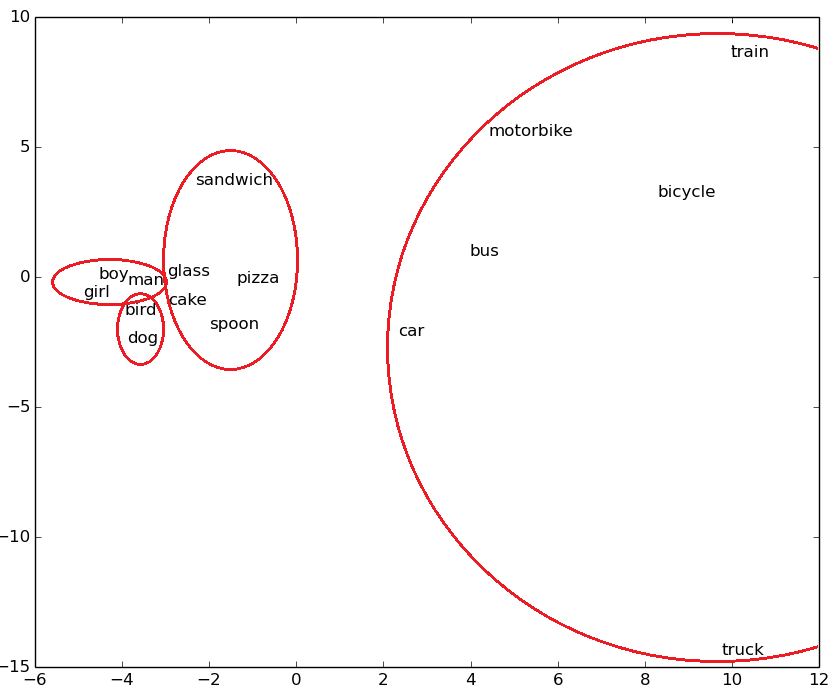}}
\caption{PCA projection on the two first eigen directions of the VIEW of words representing visual objects.}
\label{fig_word_embedding}
\end{center}
\vskip -0.3in
\end{figure}

After training, the word embedding extracted from $W_e$ enables the clustering of words into classes, as can be seen in Fig.\ref{fig_word_embedding}, which shows the PCA projection on the two first eigen directions of the embedding representing visual objects.


\begin{figure}[ht]
\vskip -0.0in
\begin{center}
\centerline{\includegraphics[width=\columnwidth]{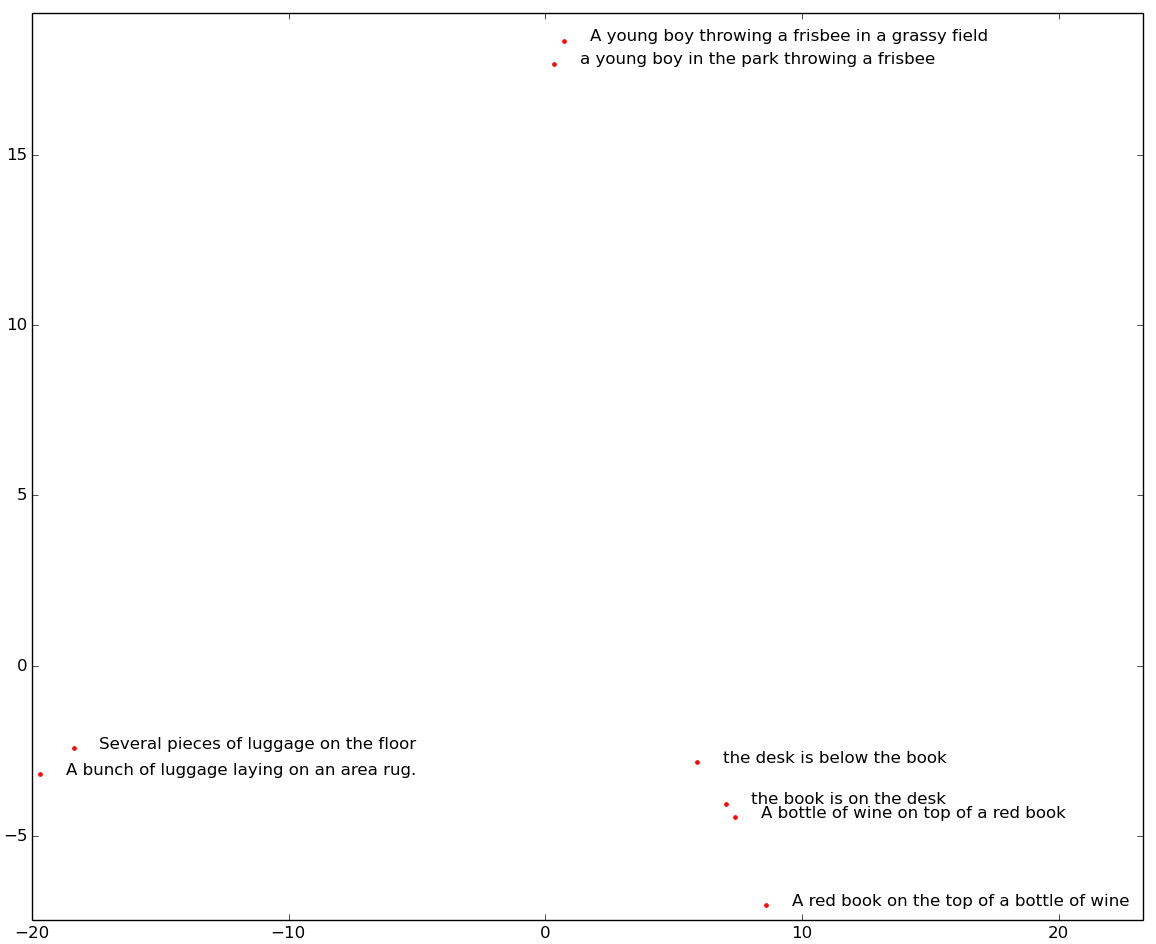}}
\caption{PCA projection of the sentence-level embedding of four pair of sentences describing the same scene in different manners.}
\label{fig_sentence_embedding}
\end{center}
\vskip -0.3in
\end{figure}

Fig.\ref{fig_sentence_embedding} shows the PCA projection of the sentence-level embedding of four pairs of sentences describing the same scene in different manners, except for a pair of sentences whose spatial meaning was changed, in order to check the sensitivity of the model to the spatial information, i.e. (``A red book on the top of a bottle of wine'', ``A bottle of wine on top of a red book''), which was plotted with a larger dispersion than the other pairs.

\subsection{Multiclass classification of word roles on SemEval}
\label{multiclass}

In this set of experiments on multiclass classification an MLP is trained to classify words into spatial roles. The adopted MLP has a single sigmoid hidden layer with 10 neurons and a linear output layer with 3 neurons, which encode the output (i.e. the predicted class: $sp$, $tr$, $lm$ or $no$ $spatial$ $role$) in one-hot vector style. The MLP receives as input the original features, $\phi_{word}(\cdot)$, extracted by the same feature function as used by the SpRL algorithm of \cite{kordjamshidi2011spatial}. These features are concatenated with features from the VIEW, in order to access the gains of using the embedding.

The MLP was trained using 15092 sentences and evaluated using 3711 sentences which compose the train and test data sets of SemEval, Task 3, SpaceEval. The results are summarized in Tables \ref{table21} and \ref{table215}.   
\begin{table}[!htbp]
\centering
\caption{Results of multiclass classification using the original features.} 
\label{table21}
\begin{tabular}{llll}
\hline
role & precision & recall & $F1$    \\
\hline
$sp$ & 0.636 & 0.839 & 0.724 \\                      
$tr$ & 0.541    & 0.723  & 0.619 \\
$lm$ & 0.405     & 0.684  & 0.509  \\
\hline
\end{tabular}
\end{table}
\begin{table}[!htbp]
\centering
\caption{Results of original features + VIEW.} 
\label{table215}
\begin{tabular}{llll}
\hline
role & precision & recall & $F1$    \\
\hline
$sp$ & 0.625  & 0.894 & 0.735 \\                      
$tr$ & 0.650    & 0.743  & 0.693 \\
$lm$ & 0.569     & 0.775  & 0.656  \\
\hline
\end{tabular}
\end{table}
The gains obtained by using VIEW are improved by selecting only 100 complementary features from the embedding employing the method explained in Section \ref{feature}, as can be seen in Table \ref{table22}.
\begin{table}[!htbp]
\centering
\caption{Results of original features + VIEW + selection complementary features.} 
\label{table22}
\begin{tabular}{llll}
\hline
role & precision & recall & $F1$    \\
\hline
$sp$ & 0.656  & 0.871 & 0.748 \\                      
$tr$ & 0.633    & 0.838  & 0.721 \\
$lm$ & 0.550     & 0.853  & 0.669  \\
\hline
\end{tabular}
\end{table}
Table \ref{table23} presents the results after the application of the fine tuning method introduced in Section \ref{max_F1} to improve the $F1$ measure.
\begin{table}[!htbp]
\centering
\caption{Results of original features + VIEW + selection complementary features + $F1$ maximization.} 
\label{table23}
\begin{tabular}{llll}
\hline
role & precision & recall & $F1$    \\
\hline
$sp$ & 0.658 & 0.869 & 0.749 \\                      
$tr$ & 0.666    & 0.813  & 0.732 \\
$lm$ & 0.600     & 0.778  & 0.678  \\
\hline
\end{tabular}
\end{table}
To compare the performance of VIEW with the usual Word2Vec embedding \cite{mikolov2013efficient}, we trained a skip-gram model\footnote{https://code.google.com/p/word2vec/} on the same COCO captions as we trained VIEW (but without visual information) and concatenated it to the original SpRL features to generate the results of Table \ref{table24}.
\begin{table}[!htbp]
\centering
\caption{Results of original features + Word2Vec embedding.} 
\label{table24}
\begin{tabular}{llll}
\hline
role & precision & recall & $F1$    \\
\hline
$sp$ & 0.612 & 0.871 & 0.719 \\                      
$tr$ & 0.602    & 0.648  & 0.624 \\
$lm$ & 0.461     & 0.552  & 0.503  \\
\hline
\end{tabular}
\end{table}
The VIEW yields performance gains in predicting all the spatial roles. These gains are improved by the application of the methods described in Sections \ref{feature} and \ref{max_F1}.

\subsection{Structured prediction of spatial triplets on SemEval}
\label{structured}

In this set of experiments on structured prediction we used the original SpRL algorithm of \cite{kordjamshidi2011spatial} not only to predict the spatial role of the words, but also to compose words into triplets $(sp,tr,lm)$, i.e. the structured output. As explained in Section \ref{task}, the algorithm uses descriptive vectors of words, $\phi_{word}(\cdot)$, and pairs of words, $\phi_{pair}(\cdot,\cdot)$. The VIEW is concatenated only to $\phi_{word}(\cdot)$, having a secondary role in this set of experiments.

The VIEW yields performance gains in classifying words into the roles $sp$ and $lm$, as can be seen in Tables \ref{table1} and \ref{table2}, which summarize the performance indexes using only the original features from $\phi_{word}(\cdot)$ and $\phi_{pair}(\cdot,\cdot)$ and using $\phi_{word}(\cdot)$ concatenated with the VIEW.
\begin{table}[!htbp]
\centering
\caption{Results of structured prediction using original features and algorithm of \cite{kordjamshidi2011spatial}.}
\label{table1}
\begin{tabular}{llll}
\hline
role/structure & precision & recall & $F1$    \\
\hline
$sp$ & 0.758 & 0.722 & 0.739 \\
$tr$ & 0.534     & 0.718  & 0.613 \\
$lm$  & 0.316     & 0.431  & 0.364 \\
$(sp,tr,lm)$ & 0.245 & 0.226 & 0.235 \\
\hline
\end{tabular}
\end{table}
\begin{table}[!htbp]
\centering
\caption{Results of original features + VIEW.}
\label{table2}
\begin{tabular}{llll}
\hline
role/structure & precision & recall & $F1$    \\
\hline
$sp$ & 0.727	& 0.757	& 0.741 \\
$tr$  & 0.513	& 0.755	& 0.611 \\
$lm$  & 0.354	& 0.555	& 0.432 \\
$(sp,tr,lm)$ & 0.228	& 0.242	& 0.235 \\
\hline
\end{tabular}
\end{table}
Table \ref{table3} shows the performance of the usual Word2Vec embedding \cite{mikolov2013efficient} concatenated to the original SpRL features, using the SpRL algorithm with the same setup assumed for the experiments with VIEW.
\begin{table}[!htbp]
\centering
\caption{Results of original features + Word2Vec embedding.}
\label{table3}
\begin{tabular}{llll}
\hline
role/structure & precision & recall & $F1$    \\
\hline
$sp$ & 0.693 & 0.726 & 0.709 \\
$tr$  & 0.496    & 0.772  & 0.604 \\
$lm$  & 0.303     & 0.462  & 0.366 \\
$(sp,tr,lm)$ & 0.170 & 0.184 & 0.177 \\
\hline
\end{tabular}
\end{table}

\section{Conclusion}
\label{conclusion}

This paper introduces a new approach in transferring spatial knowledge from multimodal data through the VIEW. The experiments provide evidence for the effectiveness of our method in transferring information useful in improving the performance of SpRL algorithms, specially in classifying words into spatial roles.

The experiments also provide evidence for the effectiveness of the algorithms for complementary feature selection and $F1$ maximization, introduced in Sections \ref{feature} and \ref{max_F1}, in improving the gains obtained by using the VIEW.

As for future work we aim at developing a method for $F1$ optimization in the structured prediction setting by extending the work \cite{Joachims2005} for the structured classification setting.

We believe that the results reported in this paper may improve with the increasing amount of annotated data. Notice that despite having a large cardinality, the COCO data set has a small variety of visual objects in its gold standard, i.e. it has only 91 object categories (including the super-categories).


\bibliography{spatial_embedding_Arxiv}
\bibliographystyle{icml2016}

\end{document}